\documentclass[a4paper, 10pt, conference]{ieeeconf}      
\usepackage{FG2021}

\FGfinalcopy 

\IEEEoverridecommandlockouts                              
\usepackage{graphicx}

\newcommand{\bfu}[1]{\textcolor{black}{#1}}

\def\FGPaperID{0003} 

\title{\LARGE \bf
The Effect of Wearing a Face Mask on Face Image Quality
}

\author{Biying Fu$^{1}$, Florian Kirchbuchner$^{1}$, Naser Damer$^{1,2}$
\\
$^{1}$Fraunhofer Institute for Computer Graphics Research IGD,
Darmstadt, Germany\\
$^{2}$Department of Computer Science, TU Darmstadt,
Darmstadt, Germany\\
Email: {biying.fu@igd.fraunhofer.de}
}

\begin{document}

\ifFGfinal
\thispagestyle{empty}
\pagestyle{empty}
\else
\author{Anonymous FG2021 submission\\ Paper ID \FGPaperID \\}
\pagestyle{plain}
\fi
\maketitle

\begin{abstract}

Due to the COVID-19 situation, face masks have become a main part of our daily life. Wearing mouth-and-nose protection has been made a mandate in many public places, to prevent the spread of the COVID-19 virus. However, face masks affect the performance of face recognition, since a large area of the face is covered. The effect of wearing a face mask on the different components of the face recognition system in a collaborative environment is a problem that is still to be fully studied. This work studies, for the first time, the effect of wearing a face mask on face image quality by utilising state-of-the-art face image quality assessment methods of different natures. 
This aims at providing better understanding on the effect of face masks on the operation of face recognition as a whole system. In addition, we further studied the effect of simulated masks on face image utility in comparison to real face masks. 
We discuss the correlation between the mask effect on face image quality and that on the face verification performance by automatic systems and human experts, indicating a consistent trend between both factors. The evaluation is conducted on the database containing (1) no-masked faces, (2) real face masks, and (3) simulated face masks, by synthetically generating digital facial masks on no-masked faces.
Finally, a visual interpretation of the face areas contributing to the quality score of a selected set of quality assessment methods is provided to give a deeper insight into the difference of network decisions in masked and non-masked faces, among other variations.

\end{abstract}

\section{Introduction}

Face recognition (FR) systems are becoming more and more widely accepted in our everyday life due to their various advantages, such as convenience \cite{krupp2013social}, contact-less capture \cite{gomez2021biometrics}, and high verification performance \cite{deng2019arcface,elasticface,pocketnet}. Due to the trend in deep-learning-based (DL-based) FR solutions, the accuracy achieved a significant improvement \cite{deng2019arcface}. This further accelerates the use of such systems in our daily life. It can be found in the automatic border control (ABC) gate to facilitate the passenger throughput, enhance process security, and reduce the workload of the human operator. It can be used to unlock personal devices like smartphones or computers. All these situations can be viewed as collaborative capture scenarios leading to high accuracy using today's FR solutions \cite{grother2018ongoing}. 

In many places, the government mandates the wearing of a face mask in public to prevent the spread of the COVID-19. However, wearing a mask can cause a large portion of the face to be obscured, affecting the performance of the FR system as studied in \cite{ngan2020ongoing,damer2021extended,damer2021masked,HomeSec}. Studies have been proposed to address the face occlusion problem, but most of them only address the occlusion in the wild, e.g., by wearing sunglasses. Only very recently, some works specifically addressed the enhancement of masked face recognition performance \cite{boutros2021mfr,boutros2021unmasking,PedroBiosig}. 

Previous studies did show the significant effect of masks on FR performance and initial solutions to enhance that. However, no previous works have investigated the effect of wearing a mask on the face image quality. Motivated by the need for a deeper understanding of the FR performance degradation caused by wearing masks, we present a systematic study on the impact of face masks on face image quality and its relation to FR performance.

To address this research gap, we conducted a set of experiments on a specifically collected database from \cite{damer2021extended} containing (1) no-masked faces, (2) real face masks, and (3) simulated face masks. We conducted a set of experiments using four of the latest proposed face image quality assessment (FIQA) approaches and investigated its relation to the face verification performance, using a COTS \cite{Neurotechnology}, one of the best performing academic FR solutions \cite{deng2019arcface}, and human expert performance.
Our experiments are designed specifically to address the following, previously unanswered, questions:
\begin{enumerate}
    \item What is the effect of wearing a mask on face image quality?
    \item Does this effect corresponds to the effect on the FR verification performance?
    \item Does the effect of the simulated mask on face image quality corresponds to that of real masks?
    \item Does the effect of the simulated mask on face image also relate to the FR verification performance, in relation to real masks?
    \item What are the face regions that attract the attention of the FIQA networks the most and how does this attention change when dealing with real or simulated (of different colors or shapes) face images?
\end{enumerate}

he paper has the following structure: in Section \ref{sec:related} the current state of research on the effect of face masks on FR is presented. In Section \ref{sec:fiqa_method}, we introduce the four FIQA methods used to calculate the face image quality of face images under different mask settings. The experimental setup is presented in Section \ref{sec:experimental_setup} including the database description, image pre-processing steps, conducted experimental setup description, and the evaluation metrics. The results and detailed experiment-driven answers to the posed research questions are stated in Section \ref{sec:results}. In addition, a visual interpretation is provided in Section \ref{sec:visualization} to illustrate the image parts that contribute the most to the face quality score on a selected set of FIQA methods in different settings. This helps us to understand and confirm the results-driven conclusions in Section \ref{sec:results}. Finally, the main take-home-messages of this work are concluded in Section \ref{sec:conclusion}.

\section{Related Work}
\label{sec:related}

Due to the current COVID-19 situation, the impact of face masks on FR systems requires more formal studies. The National Institute of Standards and Technology (NIST) has published a preliminary study result on investigating the effect of faces partially covered by protective masks on the performance of FR algorithms. The results were published in \cite{ngan2020ongoing} by evaluating 89 commercial FR solutions provided by the vendors. The results showed that even the best of FR algorithms declined substantially with masked faces. However, the study only used simulated face masks. 
The Department of Homeland Security has also conducted an evaluation with the same research scope but included more realistic data \cite{HomeSec}. They also reported significant negative effect of wearing masks on the accuracy of automatic FR. 

Damer et al.~\cite{damer2020effect,damer2021extended} investigate the effect of real and simulated masks on the FR performance by testing the verification performance on four FR solutions consisting of three academic solutions (VGGFace \cite{parkhi2015deep}, SphereFace \cite{liu2017sphereface}, and ArcFace \cite{deng2019arcface}) and one commercial solution from the vendor (MegaMatcher 12.1 SDK from Neurotechnology \cite{Neurotechnology}). A series of experiments were carried out to study the verification performance of masked faces and no-masked faces, in comparison to no-masked faces to each other. They draw the conclusion noting that a bigger effect of face masks existed on genuine pairs decision, in comparison to imposter pairs decisions. 

Recent works have also evaluated the performance of human operators in recognizing masked faces, in comparison to automatic FR solutions \cite{damer2021masked}. Consistent findings were drawn on the effect of wearing masks for both the automatic FR solutions and the human recognizers. The verification performance of human experts showed a consistent trend with that of well-performing FR solutions. Beyond the vulnerability of face recognition due to facial masks, Fang et al. \cite{fang2021real} further analysed security critical issues in terms of face presentation attacks and the detectability of these attacks when wearing a mask or when a real mask is placed on the attack. Other works focus on relating the FIQ to morphing faces \cite{9548302}, face parts \cite{9548297}, image quality \cite{DBLP:journals/corr/abs-2110-11111} or in a review \cite{DBLP:journals/corr/abs-2009-01103}.  

However, none of these works addressed the effect of wearing a mask on the face image quality and its correlation to the FR performance, which we perform extensively in this work.

\section{Face Image Quality Assessment Algorithms}
\label{sec:fiqa_method}

Unlike conventional perception quality measures, face image quality (FIQ) is viewed (and commonly intentionally designed) as a measure of the face utility to FR systems. In this paper, we selected four different FIQA methods as they demonstrated state-of-the-art performances and are based on various training and conceptualization strategies. They can be grouped into categories of either supervised, e.g., FaceQnet \cite{hernandez2019faceqnet} or unsupervised methods, e.g., rankIQ \cite{chen2015face}, MagFace \cite{meng2021magface}, and SER-FIQ \cite{terhorst2020ser}. 

The method \textbf{rankIQ} \cite{chen2015face} by Chen et al.~is a model trained to assess FIQ score based on ranked images. The main intuition derives from the premise that the quality of facial images cannot be quantified absolutely and should be considered in a relative manner. The method is trained on three different databases with varying quality face images. The training is performed in a two-staged process. Stage one solves an optimization problem by mapping the individual image features (for instance, HoG, Gabor, LBP, and CNN features) with first level rank weights. The second stage mapping uses a kernel trick to relate the learned first level individual feature score to a combined and scaled value.  

\textbf{FaceQnet} \cite{hernandez2019faceqnet} by Hernandez-Ortega et al.~is a supervised, and DL-based method using Convolutional Neural Networks to extract face embeddings. We used FaceQnet v2 which is the most recent version of \cite{hernandez2020biometric}\footnote[2]{FaceQnet Github: https://github.com/uam-biometrics/FaceQnet}. The ICAO compliance framework is used to score the images of the 300 selected individuals from the VGGFace2 gallery. For each individual, image with the highest ICAO score is used as a reference, while the non-ICAO images from the same subject are used as probes. The normalized comparison scores obtained from all the mated scores using euclidean distance are used as groundtruth quality measures of the non-ICAO probe images. FaceQnet is based on the RseNet-50 architecture \cite{he2016deep} and the fine-tuning of the successive regression layer is performed with the groundtruth score as the target label.

\textbf{MagFace} \cite{meng2021magface} by Meng et al.~is a recently developed method to generate both the face embeddings used for FR and to estimate the quality of the face image. The author proposed a loss function with adaptive margin and regularization based on its magnitude. This loss function aims at pushing the easily recognizable samples towards and the hard samples away from the class center. The face utility is thus inherently learned through this loss function during training. The magnitude of the feature vector is proportional to the cosine distance to its class center and is directly related to the face image utility. A high magnitude indicates a high face utility.  

\textbf{SER-FIQ} \cite{terhorst2020ser} proposed by Terh\"{o}rst et al.~is another unsupervised method without using any quality-labeled training images. This method relates the robustness of face representations to face quality. Randomly modified face model subnetworks are used to determine face representations. It is similar to induce noise to the face representation by using dropouts. Face images of high quality are expected to have stronger robustness and fewer variations in the face representations. In our study, we used the method finetuned with ArcFace loss \cite{deng2019arcface} using ResNet100 \cite{he2016deep} base architecture trained on MS1M database \cite{guo2016ms} named the SER-FIQ (on ArcFace) method.

\section{Experimental setup}
\label{sec:experimental_setup}
This section describes the used database in the performed experiments, the image pre-processing steps, and the evaluation metrics used to understand the addressed research questions in Section \ref{sec:results}.

\subsection{Database}
\label{sec:database}
The used database \cite{damer2021extended} serves to investigate the effect of masked faces on the face utility. The data is collected in a collaborative environment. All 48 participants were asked to take images simulating scenarios, such as the unlocking of personal devices or verification at border control. 

The participants were requested to capture the data on three different days in their residence, not required to be on three successive days. On each session, three videos (5 sec each) under different capture scenarios are collected. Selected frames without a face mask are referred to No-M setting. Selected frames with real face masks under scenario 1 are referred to as Real-M setting. The third capture scenario is to use electric lighting. To avoid the influence of electrical illumination on the quality, we do not consider this setup. The simulated mask is synthetically generated on face images from the No-M faces called Sim-M setting.

The types of real masks are not restricted and can be either standardized medical masks (such as surgical masks or FFP2 masks) or even non-standardized masks made of fabrics. The generation of digitized facial masks is based on 68 landmark detection using Dlib toolkit \cite{king2009dlib} from a face image. Based on the position of these facial landmarks, a polygon is synthesized, simulating a digital face mask of different shapes, heights, and colors. The generation of the simulated masks is according to the method described in the NIST report \cite{ngan2020ongoing} and the detailed implementation can be found on the repo \footnote[3]{Simulated Mask Github: https://github.com/fdbtrs/MFR}. For each face image, the mask type C as described in \cite{boutros2021unmasking, ngan2020ongoing} with random color is used to generate the synthetic masks in this work. 

Only the first two capture scenarios are considered in this work, as they represent the same illumination condition (no additional electric lighting). The used masked database in this paper results into the following settings: (1) 1440 faces without a mask (No-M), (2) 1440 faces with real face masks (Real-M), and (3) 1440 faces with simulated face masks (Sim-M). Samples of real face masks and generated synthetic masks of type C \cite{boutros2021unmasking} with random colors are presented in Figure \ref{fig:samples}.    
 
\subsection{Image Preprocessing}
We use the multitask cascaded convolutional networks (MTCNN) framework \cite{xiang2017joint} to detect the face in the input image. MTCNN is able to detect faces in complex and unconstrained environments. All images (masked or not) went through the same detection, crop, and alignment process as in \cite{deng2019arcface}. This is required by SER-FIQ and MagFace out of the 4 FIQA methods, and reshaped to adapt other FIQA methods to keep a comparative evaluation between these methods.

\subsection{Experiments}
To assess the effect of wearing face masks on face quality, we used the database settings as introduced in Section \ref{sec:database}. To determine the face quality, four FIQA methods (rankIQ, FaceQnet, MagFace, and SER-FIQ (on ArcFace)) are used to assess the quality metric. Each experimental setup aims at extracting findings of a certain comparison pair configuration to a specific research question:
\begin{enumerate}
    \item \textit{No-M vs Real-M} aims to investigate the effect of wearing a real face mask on face image quality. 
    \item \textit{No-M vs Sim-M} targets at investigating the effect of the simulated face masks on face image quality.
    \item \textit{Real-M vs Sim-M} further studies the effect of simulated masks in comparison to real face masks.
\end{enumerate}
The detailed research questions are answered in Section \ref{sec:results} by presenting the evaluation results followed by a discussion. 

\subsection{Evaluation metrics}

We present distributions of the estimated FIQ score for (1) No-M (blue), (2) Real-M (orange), and (3) Sim-M (green) using the 4 FIQA methods (rankIQ, FaceQnet, MagFace, and SER-FIQ (on ArcFace)) individually. \bfu{The curves are probability density functions (PDF) scaled by the number of observations across the seen quality score, such that the area under the curve sums up to 1 across the quality score.}

Furthermore, we report the Fisher Discriminant Ratio (FDR) \cite{damer2014biometric} to measure the separability of classes, i.e., between No-M vs Real-M, between No-M vs Sim-M, and between Real-M vs Sim-M. The value presents a comparative quantitative measure of the change in quality induced by the face mask. The equation for this measure is given in (\ref{eq:fdr}), 
\begin{equation}
    FDR = \frac{(\mu_1 - \mu_2)^2}{(\sigma_1)^2 + (\sigma_2)^2},
    \label{eq:fdr}
\end{equation}
where $\mu_1$ and $\mu_2$ are the scores mean value of the two considered distributions and $\sigma_1$ and $\sigma_2$ are their standard deviations. A high FDR indicates higher separability, and thus higher difference in quality. 

Besides the relative metric of FDR, we further provide the Mean-to-Mean distance metric to provide a quantitative measure between the distributions considered for each method individually. One must note that the mean-to-mean measure is not comparative between FIQA measures as they produce values on different scales, but their relative change in each method is comparable. However, the FDR is directly comparative as clarified in Equation (\ref{eq:fdr}).

To relate our findings to the verification performance and make a more quantitative analysis, we extracted verification results from Damer et al. \cite{damer2021extended}. The same experimental database is used in their work \cite{damer2021extended} and in our study. We report their reported equal error rate (EER) and the lowest false non-match rate (FNMR) for a false match rate (FMR) $\leq1.0\%$, namely the FMR100, scored on both ArcFace \cite{deng2019arcface} and one COTS FR solution \cite{Neurotechnology}. The used ArcFace version \cite{deng2019arcface} is based on ResNet-100 architecture \cite{he2016deep} pre-trained on MS-Celeb-1M database \cite{guo2016ms}. The COTS solution is the MegaMatcher 12.1 SDK from Neurotechnology \cite{Neurotechnology}. We chose to report the performance of ArcFace because it is the best performer among the three academic solutions compared in \cite{damer2021extended}, and the COTS solution achieved one of the best performance in the NIST verification competition \cite{grother2018ongoing}.

\section{Results}
\label{sec:results}
This section presents the evaluation results based on the experimental setups described in Section \ref{sec:experimental_setup}. The results and discussions are divided into answering separate research questions regarding the effect of masks on the FIQ.

\subsection{What is the effect of wearing a mask on face image quality?}
\label{subsec:real_mask}

\begin{figure*}
\centering
\begin{tabular}{cccc}
    \includegraphics[width=.22\textwidth]{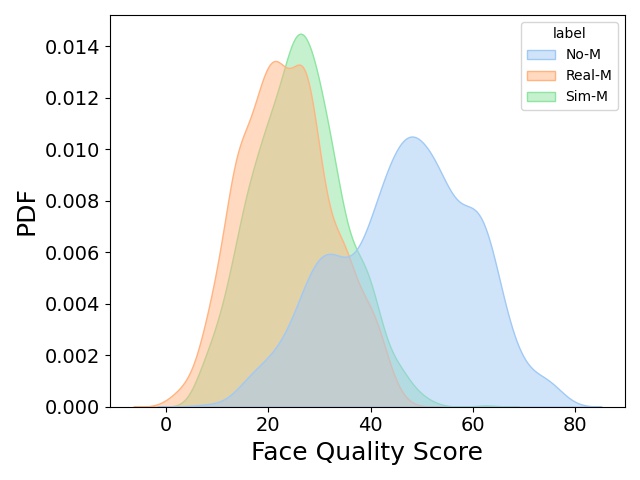} & \includegraphics[width=.22\textwidth]{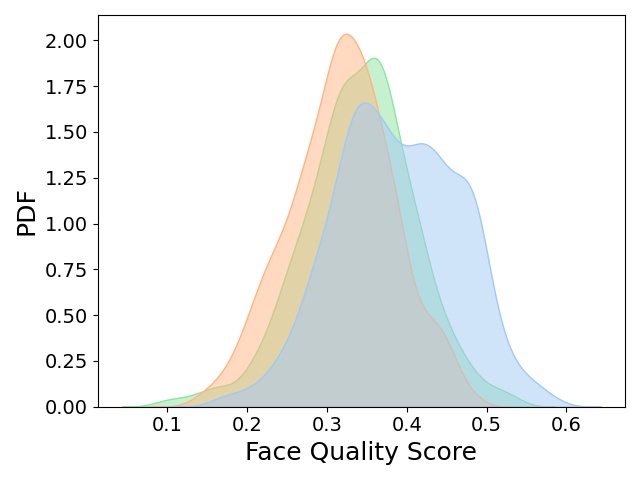} &
    \includegraphics[width=.22\textwidth]{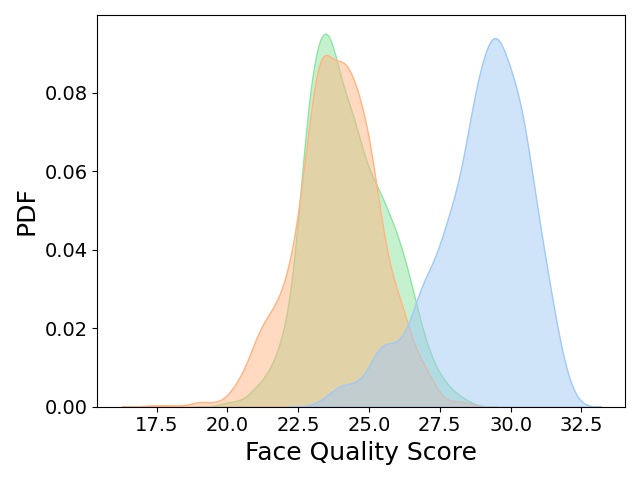} &
    \includegraphics[width=.22\textwidth]{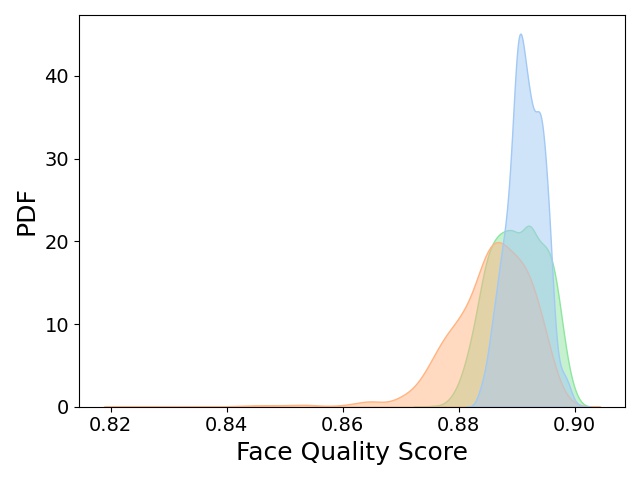} \\
     rankIQ & FaceQnet & MagFace & SER-FIQ \\[.1pt]
\end{tabular}
\caption{PDF of the quality score distribution for the settings: (1) No-M (blue), (2) Real-M (orange), and (3) Sim-M (green). The FIQ values are provided without scaling and are thus not directly comparable. For all FIQA methods, the No-M images generally provide a better FIQ score compared to masked face images. The FIQA methods (rankIQ and MagFace) show a large shift towards low face utility for masked face images, while the MagFace clearly outperforms all other methods. One can also notice a slight shift towards low quality in Sim-M in comparison to Real-M.} 
\label{fig:fiqa_methods}
\end{figure*}

Figure \ref{fig:fiqa_methods} shows the FIQ score distributions for the settings of (1) No-M (blue), (2) Real-M (orange), and (3) Sim-M (green).
Observing the results in Figure \ref{fig:fiqa_methods}, we noticed that for all FIQA methods, especially the rankIQ and MagFace, the quality score distributions for No-M setting (blue) are clearly higher compared to the masked settings. All 4 FIQA methods are understood such that, the higher the score, the higher is the face utility considered for a FR system. MagFace shows the strongest shift towards low face utility for masked images.

Table \ref{tab:mean_value} provides the quantitative measure (i.e, mean and standard deviation) of the score distributions for each of the four FIQA methods under different settings. The Mean of the No-M setting deviates from the mean of the Real-M setting in all of the FIQA methods, indicating a quality decay in masked images compared to No-M faces. In both rankIQ and MagFace we observed a measurable decay of 22.92 and 5.07 respectively in terms of mean No-M versus mean Real-M settings. In the case of SER-FIQ (on ArcFace), although less strong observable difference can be found in the mean value shift between the No-M and Real-M settings, the shape of the FIQ score distributions are different under both settings (see the increased SD in Table \ref{tab:mean_value} and Figure \ref{fig:fiqa_methods}). 

\begin{table*}
\centering
\begin{tabular}{l|l|l|l|l|l|l|l|l||l|l|l|l|}
\cline{2-13}
& \multicolumn{8}{c||}{Face Image Quality}                                                                                                                                                                                     & \multicolumn{4}{c|}{Verification Performance}                                                                                \\ 
\cline{2-13} 
& \multicolumn{2}{c|}{rankIQ \cite{chen2015face}}            & \multicolumn{2}{c|}{FaceQnet \cite{hernandez2020biometric}} & \multicolumn{2}{c|}{MagFace \cite{meng2021magface}}        & \multicolumn{2}{c||}{SER-FIQ (on ArcFace) \cite{terhorst2020ser}} & \multicolumn{2}{c|}{ArcFace \cite{deng2019arcface,damer2021extended}} & \multicolumn{2}{c|}{COTS \cite{Neurotechnology,damer2021extended}} \\ \cline{2-13} 
& \multicolumn{1}{c|}{Mean} & \multicolumn{1}{c|}{SD} & \multicolumn{1}{c|}{Mean} & \multicolumn{1}{c|}{SD} & \multicolumn{1}{c|}{Mean} & \multicolumn{1}{c|}{SD} & \multicolumn{1}{c|}{Mean}    & \multicolumn{1}{c||}{SD}    & \multicolumn{1}{c|}{EER}     & \multicolumn{1}{c|}{FMR100}     & \multicolumn{1}{c|}{EER}     & \multicolumn{1}{c|}{FMR}     \\ \hline
\multicolumn{1}{|l|}{No-M}   & 46.46                     & 13.18                   & 0.39                      & 0.08                    & 28.92                     & 1.66                    & 0.8915                       & 0.0030                     & 0\%                          & 0\%                             & 0\%                          & 0\%                          \\ \hline
\multicolumn{1}{|l|}{Real-M} & 23.54                     & 8.85                    & 0.32                      & 0.07                    & 23.85                     & 1.47                    & 0.8854                       & 0.0076                     & 2.8122\%                     & 3.3917\%                        & 1.0185\%                     & 1.0747\%                     \\ \hline
\multicolumn{1}{|l|}{Sim-M}  & 26.12                     & 8.82                    & 0.34                      & 0.07                    & 24.24                     & 1.41                    & 0.8901                       & 0.0047                     & 1.1652\%                     & 1.3002\%                        & 0.6002\%                     & 0,6337\%                     \\ \hline
\end{tabular}
\caption{The table depicts the mean quality score and the standard deviation (SD) of the quality score distribution for each individual FIQA method separated for the investigated settings of (1) No-M, (2) Real-M, and (3) Sim-M. The method \textit{rankIQ} and \textit{MagFace} show the largest deviation in terms of the mean quality score between the No-M setting and the masked setting. Extracted from \cite{damer2021extended}, we also provided the EER and FMR100 for an academic FR system (ArcFace \cite{deng2019arcface}) and a COTS FR system \cite{Neurotechnology}. The verification used No-M reference image and compare it to different probes (No-M, Real-M, and Sim-M).}
\label{tab:mean_value}
\end{table*}

Table \ref{tab:fdr_value} provides the FDR measure and the Mean-to-Mean distance indicating the quality separability (shift) between the different mask settings. Looking at the results in Table \ref{tab:fdr_value}, we observed the MagFace providing the strongest shift towards low face utility for Real-M setting in terms of the cross-metric comparable FDR value with a value of 5.21 (FDR No-M vs the Real-M). This finding is in agreement with the observations made in Figure \ref{fig:fiqa_methods}.

\begin{table*}
\centering
\begin{tabular}{l|l|l|l|l|}
\cline{2-5}
& \multicolumn{1}{c|}{rankIQ \cite{chen2015face}} & \multicolumn{1}{c|}{FaceQnet \cite{hernandez2020biometric}} & \multicolumn{1}{c|}{\textbf{MagFace} \cite{meng2021magface}} & \multicolumn{1}{c|}{SER-FIQ (on ArcFace) \cite{terhorst2020ser}} \\ \hline
\multicolumn{1}{|l|}{FDR No-M vs Real-M}        & 2.08                                 & 0.46                          & \textbf{5.21}                         & 0.56                         \\ \hline
\multicolumn{1}{|l|}{FDR No-M vs Sim-M}         & 1.64                                 & 0.21                          & \textbf{4.62}                         & 0.06                         \\ \hline
\multicolumn{1}{|l|}{FDR Real-M vs Sim-M}       & 0.04                                 & 0.05                          & 0.04                                  & 0.28                         \\ \hline
\hline
\multicolumn{1}{|l|}{Mean-Mean No-M vs Real-M}  & 22.92                       & 0.07                          & 5.07                                  & 0.0061                       \\ \hline
\multicolumn{1}{|l|}{Mean-Mean No-M vs Sim-M}   & 20.34                       & 0.05                          & 4.68                                  & 0.0014                       \\ \hline
\multicolumn{1}{|l|}{Mean-Mean Real-M vs Sim-M} & 2.58                        & 0.02                          & 0.39                                  & 0.0047                       \\ \hline
\end{tabular}
\caption{The FDR (according to Equation \ref{eq:fdr}) value and the Mean-to-Mean distance (Mean-Mean) for (1) No-M vs Real-M, (2)  No-M vs Sim-M, and (3)  Real-M vs Sim-M values are provided for all 4 FIQA methods investigated. \textit{MagFace} shows the highest decay in face utility for the masked setting. Though it is much complicated to discriminate the decay in utility between real mask and simulated mask, which corresponds to the findings in \cite{damer2021extended}. }
\label{tab:fdr_value}
\end{table*}

\begin{figure}
\centering
\begin{tabular}{c}
    \includegraphics[width=.45\textwidth]{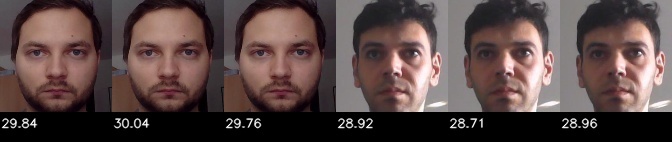} \\
    MagFace metric on faces with no mask\\[.1pt]
    \includegraphics[width=.45\textwidth]{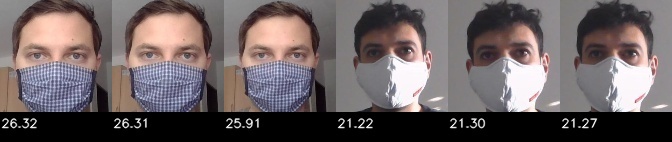} \\
    MagFace metric on faces with real mask\\[.1pt]
    \includegraphics[width=.45\textwidth]{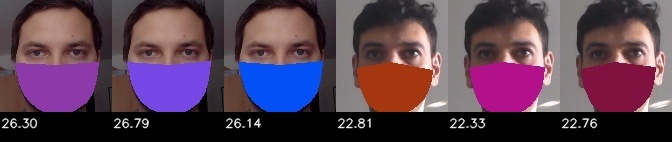} \\
    MagFace metric on faces with simulated mask\\[.1pt]
\end{tabular}
\caption{Sample images showing No-M faces, Real-M and Sim-M face images. The quality score calculated by MagFace \cite{meng2021magface} is illustrated below the image. A high utility score indicates a good sample quality. The score degrades with the increased coverage of the face as it is observed in case of real masks, even when other factors such as inconsistent illumination (three image to the right) have already degraded the quality score.} 
\label{fig:samples}
\end{figure}

These findings are reflected in the sample images depicted in Figure \ref{fig:samples}. In Figure \ref{fig:samples} we showed sample images with the FIQ score using the MagFace method \cite{meng2021magface}. Comparing sample images from first and second row, a large part of the face area is covered by a real mask. One noticed that strong light reflection does affect the quality significantly. However, the drop of quality caused by wearing a mask still persists. Similar quality decay is observed for simulated masks generated on the same No-M images.

To sum up, there is a clear indication that wearing a mask does cause the measured FIQ to drop across all the considered state-of-the-art FIQA measures. This drop is relatively stronger for rankIQ and MagFace approaches.

\subsection{Does this effect (of masks on FIQ) corresponds to the effect on the FR verification performance?}

As we are using the same database introduced in \cite{damer2021extended}, we correlate our results to the verification results presented for the two top performing FR solutions in \cite{damer2021extended} as described in Section \ref{sec:experimental_setup}.

The effect of the mask on the verification performance is practically related to the scenario where the reference is No-M and the probe is masked as indicated in \cite{damer2021extended, ngan2020ongoing}. This scenario corresponds to, for example, the case in automated border control where the reference image stored on the passport is not masked, but the live image captured at the border gate can be masked. Therefore, to correlate our quality investigation to the verification performance, we assumed to have always a no mask reference and a probe that is either No-M, Real-M, or Sim-M, and therefore we point out the experiment by the type of probe data used. The verification results achieved by the COTS and ArcFace solutions described in Section \ref{sec:experimental_setup} are presented in Table \ref{tab:mean_value}.
The observed effect of masks on FIQ is consistent with the effect observed on the verification performance. Error-free verification performance was achieved for No-M face setting. However, the verification performance decreased in the Real-M face setting, as seen in Table \ref{tab:mean_value}. The EER is increased from error-free achieved by No-M setting to 2.8122\% achieved by Real-M setting for ArcFace and to 1.0185\% for the COTS system. The same is observed for FMR100 with an increased error rate to 3.3917\% for ArcFace and to 1.0747\%, for COTS respectively, which goes consistently with the decay in the face image qualities. 

Utilizing the same database, the work in \cite{damer2021masked} further compared the Human performance on the masked face verification. \bfu{The same setups (No-M, Real-M, and Sim-M) that are evaluated by automatic FR solutions are evaluated by human experts in \cite{damer2021masked}. Two different crop styles, crop-1 and crop-2 are applied to the face images presented to the human observer. These crop styles should exclude larger changes in the hairstyle, background, or even hairlines. Both crop styles are based on detected faces using MTCNN, while crop-1 removed 10\% from all borders, and crop-2 removed only 5\%. Similar to software solutions, the performance of human experts also drops notably in terms of EER and FMR100 in the Real-M setting compared to No-M setting. In Real-M face setting, the EER increased from error-free verification (in the No-M setting) to 2.64\% and 0.34\% for crop-1 and crop-2 respectively. The same is observed in terms of FMR100 measure. In Real-M setup, the FMR increased from 0.00\% to 4.00\% for crop-1. All the human expert results are reported as in \cite{damer2021masked}. It is noted in the results for crop-2, that a wider crop around the face region seems to improve the verification performance significantly indicating the experts' decisions might be influenced by properties like hairstyle, hair colors, and the forehead area.}

To sum up, a clear correspondence is noticed between the effect of real masks on both the face image quality (using different FIQA) and the face verification performance by different FR solutions and by human experts.

\subsection{Does the effect of the simulated mask on face image quality corresponds to that of real masks?}

To answer this posed question, we start by viewing the Figure \ref{fig:fiqa_methods}. The face quality distribution of Sim-M (green) strongly overlaps with the face quality distribution of Real-M (orange) in all 4 FIQA methods. A strong left shift away from the No-M (blue) distribution can be observed for rankIQ and MagFace which indicates a drop in Sim-M utilities and corresponds to the results for Real-M faces as discussed in Section \ref{subsec:real_mask}. 

Similar findings can be drawn from Table \ref{tab:mean_value}. For all 4 FIQA methods, a clear decrease in the mean value can be observed in terms of face utility. In both rankIQ and MagFace, we observed a measurable decay of 20.34 and 4.68 respectively, in terms of Mean-Mean No-M vs Sim-M settings. The same result is confirmed in terms of the FDR of No-M vs. Sim-M in Table \ref{tab:fdr_value}, with a FDR=4.62 for MagFace and a FDR=1.64 for rankIQ. In addition, it should be noted that the shift of the face utility of the Sim-M setting is closer to the No-M setting compared to the Real-M setting. In the case of MagFace, a smaller mean-shift of 4.68 is observed for Mean-Mean No-M vs Sim-M compared to 5.07 for Mean-Mean No-M vs Real-M. For rankIQ the corresponding number is increased from 20.34 to 22.92 moving from Mean-Mean No-M vs Sim-M to Mean-Mean No-M vs Real-M respectively. The same goes for the other FIQA methods.

To sum up, such as real masks, simulated masks do decrease the FIQ using all the considered FIQA methods. However, our experiment showed that this decay in quality is lower in the case of simulated masks in comparison to real masks. This might be related to the higher textural information and interaction with the environment (illumination) in the real mask scenario.
Despite the possibility of creating much more diverse simulated masks, our results point out the difficulty in creating a fully representative evaluation based on real masks.

\subsection{Does the effect of the simulated mask on face image also relate to the FR verification performance, in relation to real masks?}

In terms of the verification performance, as presented in Table \ref{tab:mean_value}, we consider the case where the reference is No-M and the probe is Sim-M face images. In comparison to the No-M face settings, there is an increase in EER from 0.00\% to 1.16\% for ArcFace FR solution and from 0.00\% to 0.60\% for the COTS \cite{Neurotechnology} FR solution. A similar increase for FMR100 is observed for ArcFace (from 0.00\% to 1.30\%) and COTS (from 0.00\% to 0.6337\%) respectively in the same Sim-M face setting.

However, this verification error induced by the Sim-M faces is lower than that induced by Real-M faces. In terms of EER, a relative decrease of 58.57\% and 41.07\% is observed for ArcFace and COTS respectively when moving from Real-M to Sim-M setting. Consistently to the effect on the verification performance, the FIQ follows a similar trend. The Sim-M faces show higher FIQ on all considered measures when compared to Real-M faces. The distributions of the quality measures of the real and simulated masks strongly overlap, however, with an observable slight shift to higher quality range for the simulated masks, as seen in Figure \ref{fig:fiqa_methods}.

In further agreement to the verification decision of human experts, \cite{damer2021masked} found that it is less challenging to compare a no-masked reference to a simulated masked probe (Sim-M) than to real masked probe (Real-M) images. \bfu{The EER in the Sim-M face setting is improved to 1.6\% from 2.6\% in the Real-M face setting for crop-1, while even larger improvement is observed for crop-2 from 0.3\% (Real-M) to 0.0\% (Sim-M).}

To sum up, consistent with the FIQ, both the verification performance and FIQ decrease when dealing with simulated masks (Sim-M), however, for both cases, in a less degree than real masks, again showing a clear correlation between the measured qualities and the verification performance. This result is also consistent with both the automated FR solutions and the human experts decisions.

\section{Visualization}
\label{sec:visualization}
In addition, to evaluate the effect of (1) No-M, (2) Real-M, and (3) Sim-M on FIQ, we further explore the image regions that contribute the most to their performance on a selected set of quality assessment methods for different settings. We use the Score-CAM \cite{wang2020score} method to display the attention area of the network's decision. Score-CAM is designed for displaying visual explanations for convolutional neural networks (CNNs). Unlike other network visualization approaches, such as Gradient visualization \cite{DBLP:journals/corr/SimonyanZ14a}, Perturbation \cite{DBLP:conf/kdd/Ribeiro0G16}, and Class Activation Map (CAM) \cite{DBLP:conf/cvpr/ZhouKLOT16} which are based on the gradients of the network's backward passes, Score-CAM can be applied only on the forward pass by combining the weighted activation maps. More importantly, as Wang et al. demonstrated \cite{wang2020score},  Score-CAM achieved better visual performance and fairness for interpreting the decision making process.
FaceQnet and MagFace are visualized here using their ResNet-x base architecture. This visualization is unfortunately not applicable to both, the rankIQ due to the fusion of selected features in the two-stage process, and the SER-FIQ with the multiple dropout runs applied to modified network architectures.

\begin{figure}
\centering
\begin{tabular}{c}
    \includegraphics[width=.43\textwidth]{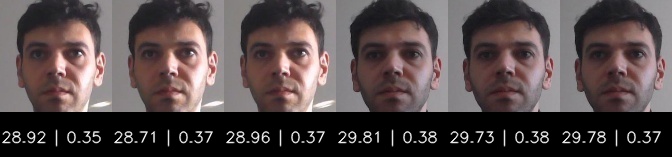} \\
    No-M Faces\\[.1pt]
    \includegraphics[width=.43\textwidth]{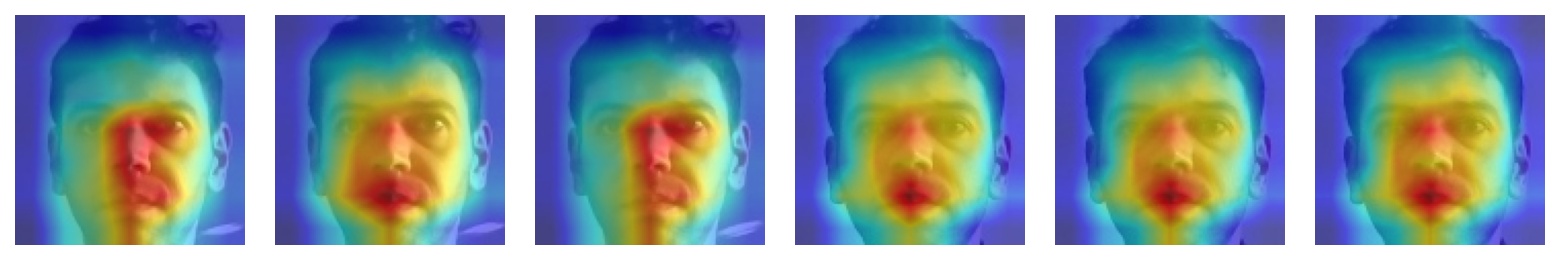} \\
    Visualization MagFace Attention\\[.1pt]
    \includegraphics[width=.43\textwidth]{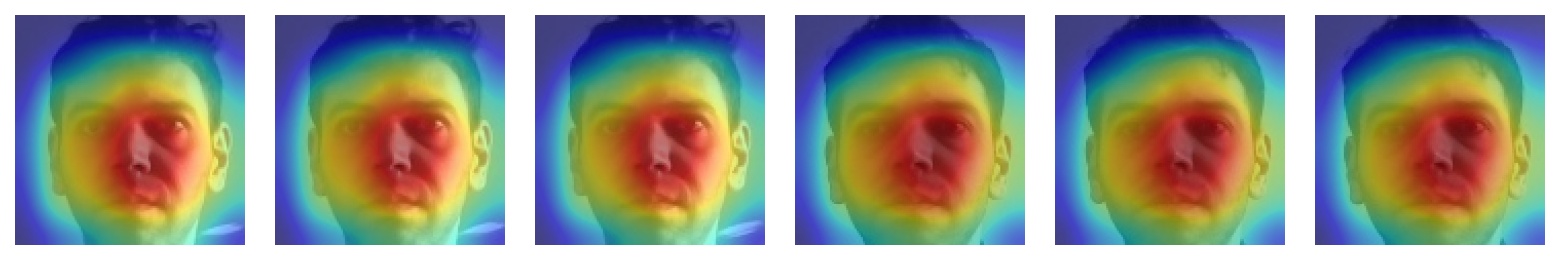} \\
    Visualization FaceQnet Attention\\[.1pt]
\end{tabular}
\caption{Score-Cam visualization of faces of the No-M setting. The attention area of the MagFace method is more locally refined compared to FaceQnet. The attention in both cases is on the center of the face, including the eyes, nose, and mouth regions. The quality scores for each image are provided in the Form of (MagFace$|$FaceQnet).} 
\label{fig:vis_no_mask}
\end{figure}

Figure \ref{fig:vis_no_mask} illustrates the visualization of the attention of the FIQA networks (MagFace and FaceQnet) on No-M images from different capture sessions. The images are intentionally selected to represent different illumination cases (illumination change is not instructed to the subject).
The main attention of these FIQA algorithms is centered on the central facial region and gradually spreads to the edges of the facial region with attenuating intensity. The attention area is more homogeneous for FaceQnet compared to MagFace specifying that the MagFace is more locally refined. In general, the highest intensity regions include the eyes, nose, and mouth.

\begin{figure}
\centering
\begin{tabular}{c}
    \includegraphics[width=.43\textwidth]{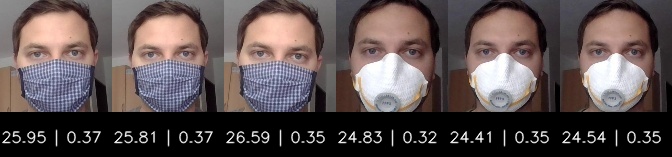} \\
    Real-M faces\\[.1pt]
    \includegraphics[width=.43\textwidth]{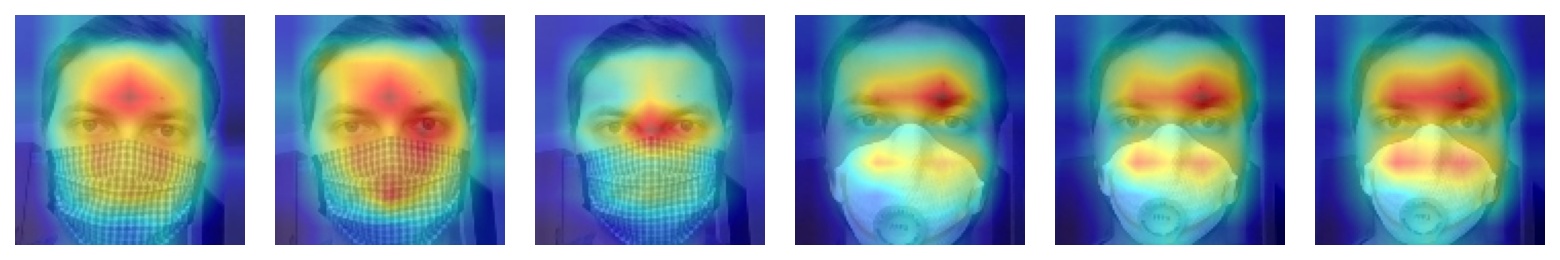} \\
    Visualization MagFace Attention\\[.1pt]
    \includegraphics[width=.43\textwidth]{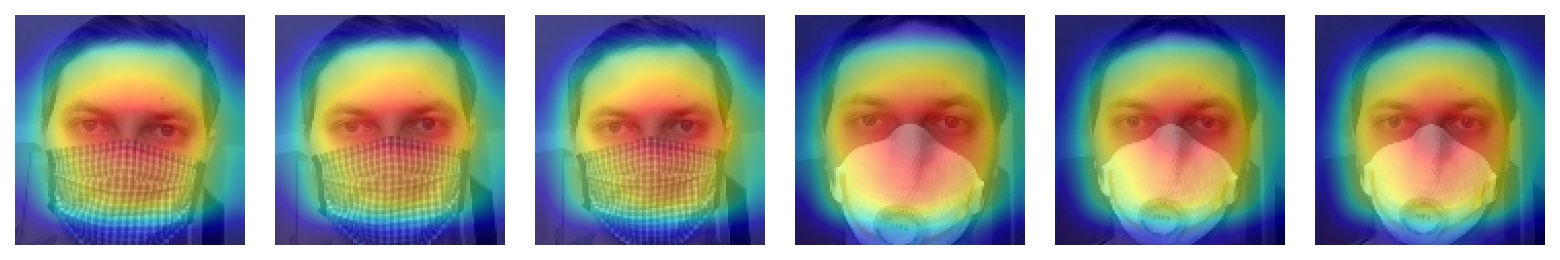} \\
    Visualization FaceQnet Attention\\[.1pt]
\end{tabular}
\caption{Score-Cam \cite{wang2020score} visualization of faces with real face masks. The focus of both quality estimation solutions is focused on the upper face region and looses focus in the covered areas of the nose and mouth, in comparison to No-M faces in Figure \ref{fig:vis_no_mask}. The scores are provided in the Form of (MagFace$|$FaceQnet).}
\label{fig:vis_real_mask}
\end{figure}

In Figure \ref{fig:vis_real_mask}, we illustrated images with Real-M faces from two different sessions with two different types of real masks. In Figure \ref{fig:vis_real_mask}, and for both the MagFace and FaceQnet, it is observable that the high attention area is significantly smaller in comparison to the No-M images in Figure \ref{fig:vis_no_mask}.
This reduction is mainly in the area covered by the mask. With the mask, the attention of the FIQA networks is centered on the upper face (eyes region) while the network have less focus on the covered nose and mouth regions. It is also noted, that in some cases, the mask seems to confuse the FIQA network and shift the attention to an area above the eyes.

\begin{figure}
\centering
\begin{tabular}{c}
    \includegraphics[width=.47\textwidth]{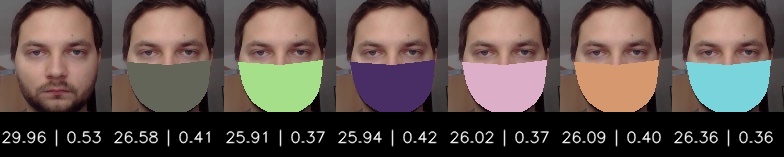} \\
    Mask Type C as defined in \cite{boutros2021unmasking, ngan2020ongoing} with random colors\\[.1pt]
     \includegraphics[width=.47\textwidth]{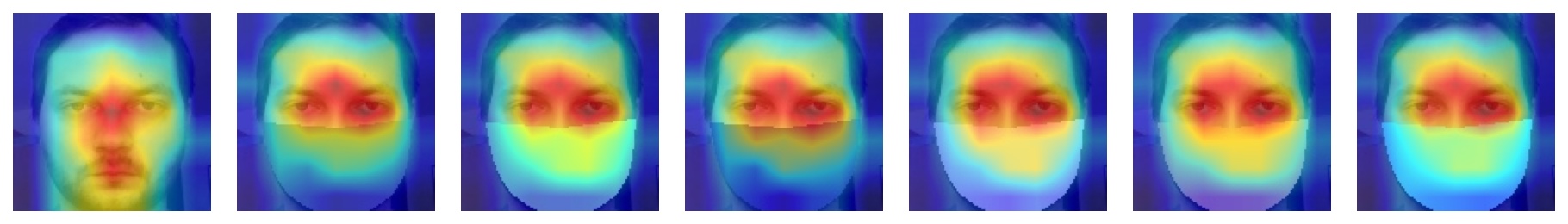} \\
    Attention Maps of MagFace\\[.1pt]
    \includegraphics[width=.47\textwidth]{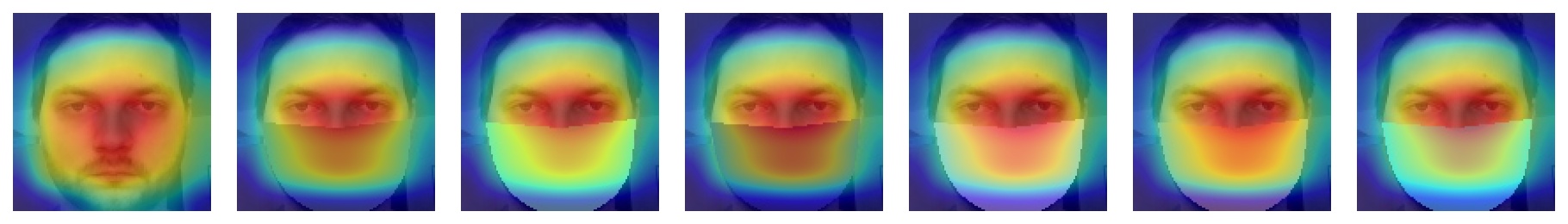} \\
    Attention Maps of FaceQnet\\[.1pt]
\end{tabular}
\caption{Simulated mask is applied to the same face image. \bfu{First, the type C mask in \cite{boutros2021unmasking, ngan2020ongoing} with random color are applied to the No-M face.} Score-Cam \cite{wang2020score} visualization of the same face with type C mask of random colors are shown for FaceQnet and MagFace below. Both FaceQnet and MagFace react slightly differently on identical masks with different colors. The scores are provided in the Form of (MagFace$|$FaceQnet).} 
\label{fig:vis_sim_mask_color}
\end{figure}

To visualize the effect of simulated masks, we generated different colored and typed masks on the same No-M face image. First, we applied Type C masks as in \cite{ngan2020ongoing} using random colors to the same No-M face image. This type of simulated mask is used in our experimental study. In Figure \ref{fig:vis_sim_mask_color} (a), one notices that the color of the mask, even on the same image, the same mask shape, and boundaries, causes changes in the networks' decision. This can be due to the nature of the neural network operations, where series of multiplications are effected by the initial pixel values. This is also reflected in the slightly different attention regions in the images in Figure \ref{fig:vis_sim_mask_color} when having different mask colors.

\begin{figure}
\centering
\begin{tabular}{c}
    \includegraphics[width=.47\textwidth]{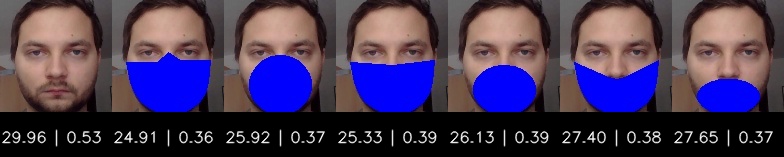} \\
    Mask Type A, B, C, D, E, and F as defined in \cite{boutros2021unmasking, ngan2020ongoing} \\[.1pt]
     \includegraphics[width=.47\textwidth]{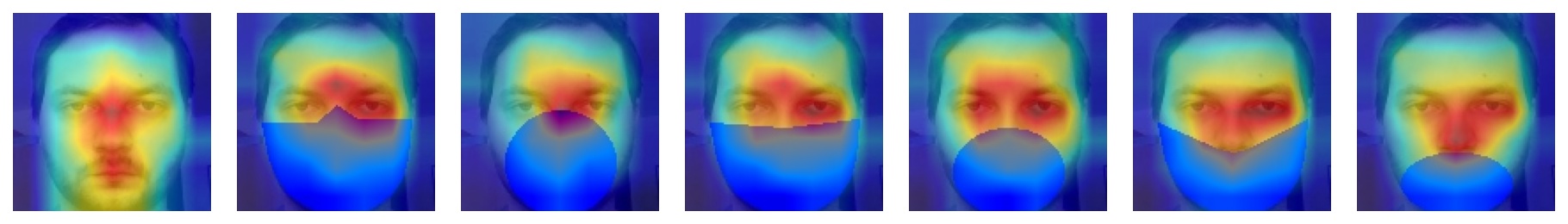} \\
    Attention Maps of MagFace\\[.1pt]
    \includegraphics[width=.47\textwidth]{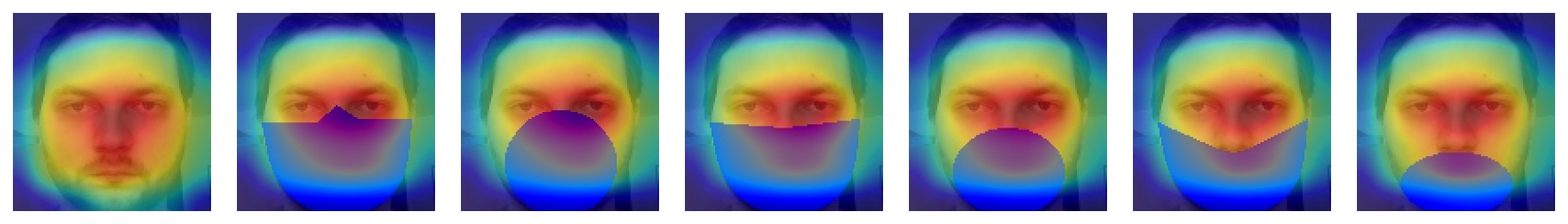} \\
    Attention Maps of FaceQnet\\[.1pt]
\end{tabular}
\caption{Here, we applied the different mask types A to F as defined in \cite{boutros2021unmasking, ngan2020ongoing} with the same color to the same No-M face. The higher coverage reduces the attention on the nose region. Especially for the MagFace, a strong attention is noticed on the upper boundary of the simulated mask. The scores are provided in the Form of (MagFace$|$FaceQnet).} 
\label{fig:vis_sim_mask_types}
\end{figure}

Additionally, we applied the different mask types A to F as defined in \cite{boutros2021unmasking, ngan2020ongoing} with same color to the same No-M face. From Figure \ref{fig:vis_sim_mask_types}, it is noted that both MagFace and FaceQnet have high attention on the upper face region for most images (as for the real masks). Figure \ref{fig:vis_sim_mask_types} simulated the difference in face coverage by applying different mask types to the same image. It is observed that higher coverage of the simulated mask reduces the attention on the nose region, especially for the MagFace solution. Moreover, the higher coverage of the face by simulated masks causes the estimated FIQ to drop, although more clearly in the MagFace FIQA.

Finally we compared the networks behaviour towards real masks and simulated masks. Looking at Figure \ref{fig:vis_real_mask}, networks clearly focus on the visible regions of the face while relatively neglects the regions covered by the real mask. However, they also pay attention to the masked areas where a clear nose structure is visible. This intuitive behaviours is not completely reflected in the networks behaviours towards simulated masks in Figure \ref{fig:vis_sim_mask_color} and Figure \ref{fig:vis_sim_mask_types}. The main attention seems to be only on the visible face area. This might be caused by the nature of the simulated mask texture that does not possess the expected natural image texture distribution caused by the nose shape.

To sum up, the FIQA networks seem to loose dependency on the covered areas in the masked images in an intuitive manner. \bfu{This is only partially true for simulated masks (in comparison to real masks) where the networks' attention, in some cases, are affected by the mask color, which might be due to the nature of the neural network operations.}

\section{Conclusion}
\label{sec:conclusion}
To analyse the effect of wearing a face mask on the FIQ, we conducted extended experiments on a collaborative and consistent database \cite{damer2021extended} containing (1) No-M faces, (2) Real-M faces, and (3) Sim-M faces using four state-of-the-art FIQA methods.
We studied the effect on face quality when wearing real-world masks in comparison to faces without a mask and found that wearing a face mask reduces the face utility for all investigated FIQA methods. The finding in terms of face utility drop for masked images also agrees with the effect of face masks on face verification performance by automatic FR systems and human experts. The findings were both consistent for real and simulated masks. This result follows the fact that FIQA methods are designed to the task of measuring the utility of a face image for FR tasks. We further found that the effect on FIQ caused by simulated mask does not simply reflect that of a real mask, also in agreement with the verification results for automated FR solutions and human experts decisions in \cite{damer2021masked}. 

Based on the visualization of the networks attention area, we expected the face utility to decrease for masked images due to the loose dependence on the covered face areas. \bfu{It is also interesting to note that for simulated masks, the networks' attention, in some cases, are also affected by the mask color, which might be due to the inherent nature of the neural network operations.}


{\small
\bibliographystyle{ieee}
\bibliography{egbib}
}

\end{document}